\documentclass[letterpaper, 10 pt, conference]{ieeeconf}

\IEEEoverridecommandlockouts                              
\usepackage{amsmath,mathtools} 
\usepackage{amssymb}  
\usepackage{graphicx}
\usepackage[hidelinks]{hyperref}
\usepackage{siunitx}
\usepackage{subcaption}
\usepackage{booktabs}
\usepackage{xcolor}
\usepackage{textcomp}
\usepackage{gensymb} 
\usepackage{multirow} 
\usepackage{comment}
\usepackage{verbatim}
\usepackage[skip=0pt, font=footnotesize]{caption}
\usepackage{mathtools}
\usepackage{bm}
\usepackage[noadjust]{cite}
\usepackage{makecell}
\usepackage{tabularx}
\usepackage{pifont}
\usepackage{layouts}
\usepackage{physics}
\usepackage{algorithm}
\usepackage{algpseudocode}
\usepackage{layouts}
\usepackage{lipsum}

\IEEEaftertitletext{\vspace{-0.5\baselineskip}}

\makeatletter
\renewcommand{\@IEEEsectpunct}{ \,}
\makeatother

\definecolor{peru}{rgb}{0.803921568627451, 0.5215686274509804, 0.24705882352941178}
\definecolor{violet}{rgb}{0.9333333333333333, 0.5098039215686274, 0.9333333333333333}
\definecolor{greeN}{rgb}{0.17254901960784313, 0.6274509803921569, 0.17254901960784313}
\definecolor{stage0}{RGB}{187,248,255}
\definecolor{stage1}{RGB}{250,255,187}
\definecolor{stage2}{RGB}{187,255,196}
\definecolor{stage0_dark}{RGB}{0,180,200}
\definecolor{stage1_dark}{RGB}{200,180,0}
\definecolor{stage2_dark}{RGB}{0,200,0}

\definecolor{centerline}{RGB}{51,51,255}
\definecolor{exterior}{RGB}{255,153,51}
\definecolor{interior}{RGB}{0,153,0}
\definecolor{slalom}{RGB}{255,51,51}

\def\anonymous{1} 

\DeclareRobustCommand{\rchi}{{\mathpalette\irchi\relax}}
\newcommand{\irchi}[2]{\raisebox{\depth}{$#1\chi$}} 
\newcommand{\txi}{{}_t\xi}
\newcommand\ringring[1]{%
  {
   \mathop{\kern0pt #1}\limits^{
     \vbox to-1.85ex{
       \kern-2ex 
       \hbox to 0pt{\hss\normalfont\kern.1em \r{}\kern-.45em \r{}\hss}%
       \vss 
     }
   }
  }
}

\DeclareRobustCommand{\bi}{\textbf{i}}
\DeclareRobustCommand{\bj}{\textbf{j}}
\DeclareRobustCommand{\bz}{\textbf{k}}

\newcolumntype{M}[1]{>{\centering\arraybackslash}m{#1}}

\makeatletter
\def\endthebibliography{%
	\def\@noitemerr{\@latex@warning{Empty `thebibliography' environment}}%
	\endlist
}
\makeatother

\IEEEoverridecommandlockouts
\usepackage{tikz}
\usepackage{textcomp}
\usepackage{hyperref}
\usepackage{lipsum}

\newcommand\copyrighttext{%
    \footnotesize Published in IEEE/RSJ International Conference on Intelligent Robots and Systems (IROS), Detroit, USA, 2023.\newline
    \textcopyright 2023 IEEE. Personal use of this material is permitted.
	Permission from IEEE must be obtained for all other uses, in any current or future media, including reprinting/republishing this material for advertising or promotional purposes, creating new collective works, for resale or redistribution to servers or lists, or reuse of any copyrighted component of this work in other works.
 }
\newcommand\copyrightnotice{%
	\begin{tikzpicture}[remember picture,overlay]
		\node[anchor=south,yshift=10pt] at (current page.south) {\fbox{\parbox{\dimexpr\textwidth-\fboxsep-\fboxrule\relax}{\copyrighttext}}};
	\end{tikzpicture}%
}

\title{\LARGE \bf
SCTOMP: \\Spatially Constrained Time-Optimal Motion Planning}
\if\anonymous1
\author{Jon Arrizabalaga$^{1}$ and Markus Ryll$^{1,2}$
	\thanks{$^{1}$Autonomous Aerial Systems, School of Engineering and Design,  Technical University of Munich, Germany. E-mail: {\tt\small jon.arrizabalaga@tum.de} and {\tt\small markus.ryll@tum.de}}%
	\thanks{$^{2}$Munich Institute of Robotics and Machine Intelligence (MIRMI), Technical University of Munich}
}
\fi
\begin{document}

\maketitle
\copyrightnotice
\begin{abstract}
This work focuses on spatial time-optimal motion planning, a generalization of the exact time-optimal path following problem that allows a system to plan within a predefined space. In contrast to state-of-the-art methods, we drop the assumption of a given collision-free geometric reference. Instead, we present a three-stage motion planning method that solely relies on start and goal locations and a geometric representation of the environment to compute a time-optimal trajectory that is compliant with system dynamics and constraints. The proposed scheme first finds collision-free navigation corridors, second computes an obstacle-free Pythagorean Hodograph parametric spline along each corridor, and third, solves a spatially reformulated minimum-time optimization problem at each of these corridors. The spline obtained in the second stage is not a geometric reference, but an extension of the free space associated with its corridor, and thus, time-optimality of the solution is guaranteed. The validity of the proposed approach is demonstrated by a well-established planar example and benchmarked in a spatial system against state-of-the-art methodologies across a wide range of scenarios in highly congested environments.

\end{abstract}
\vspace{1mm}
\textbf{Video}: \url{https://youtu.be/zGExvnUEfOY}
\section{INTRODUCTION}
\noindent Time-optimal motion planning within cluttered environments poses multiple challenges. The underlying motion planning scheme needs to compute a set of input commands that drive the system from its current state to a goal location in minimum-time, without compromising system constraints and spatial bounds. Thus, its solution implies a trading-off between time-optimality and spatial-awareness. 

The de facto approach to solve this problem has been to decouple it into two stages \cite{tranzatto2022cerberus,foehn2022alphapilot,hanover2023past,arrizabalaga2021caster}. First, the \emph{path planning} stage determines a collision-free geometric path according to high level -- task related -- commands \cite{gasparetto2015path}. Second, the predefined path is (exactly) tracked either by \emph{path tracking} or \emph{path following}. The former computes a dynamically feasible timing law for traversing along the predetermined geometric path -- \emph{when} to be \emph{where} --, while the latter introduces the timing law and the (bounded) distance to the path as control freedoms \cite{verscheure2009time,faulwasser2015nonlinear,lam2010model}.

When tackling time-optimality, previous work mainly focused on the second stage. \emph{Time-optimal path-tracking} for robotic manipulators is a long studied problem \cite{shiller1989robot,shin1985minimum,bobrow1985time}. Its convexity was proven in \cite{verscheure2009time} and in \cite{lipp2014minimum} it was extended to a wider range of systems. Enhanced numerical implementations to exploit this convexity were presented in \cite{pham2018new,consolini2019optimal}. Regarding \emph{time-optimal path-following}, \cite{verschueren2016time,spedicato2017minimum} leveraged a spatial reformulation of the system dynamics, allowing to compute minimum-time trajectories \emph{around} the predefined path. Similarly, assuming waypoints to be predetermined, \cite{foehn2021time} made use of Complementary Progress Constraints (CPC) to compute time-optimal trajectories for quadrotors. Lastly, exploiting optimal control, nonlinear model predictive control (NMPC) methods based on the aforementioned spatial reformulation \cite{arrizabalaga2022towards,kloeser2020nmpc} and contouring control \cite{liniger2015optimization,romero2022model} have shown to approximate time-optimal performance in race-alike scenarios, i.e., \emph{around} a predefined geometric path (the centerline of the track). However, the optimality of all these approaches is upper bounded by the geometric reference predetermined by the path planning stage. In other words, only if \emph{both} the path planning and path tracking/following stages are optimal, will the resultant planned trajectory be optimal, implying that the decoupled nature of these methods jeopardizes the time-optimality of the planned trajectories. 

To overcome this conceptual shortcoming, \cite{penicka2022minimum} presented a hierarchical sampling-based method capable of computing time-optimal trajectories to fly a quadrotor over a set of waypoints within cluttered environments. This was further extended in \cite{penicka2022learning}, where planning and control were simultaneously solved with a deep Reinforcement Learning (deep RL) approach. Despite the ability to roughly approximate time-optimal trajectories in congested scenarios, none of these methods can guarantee that the resultant trajectory is the true-optimal. On the one hand, the sampling-based nature of \cite{penicka2022minimum} renders it nondeterministic, introducing randomness into the solution. On the other hand, the additional tracking capabilities brought by the end-to-end learning approach in \cite{penicka2022learning} come at the expense of 1) system-specificity -- changes in the system dynamics require retraining -- and 2) sub-optimal trajectories resulting from learning the trade-off between flying safe and fast. 

This raises the question on how to formulate a motion planning scheme, applicable to any system and constrained environment, that guarantees to compute the time-optimal trajectory compliant with system dynamics, state/input constraints and spatial bounds. To answer this question, in this paper we present a \textbf{S}patially \textbf{C}onstrained \textbf{T}ime-\textbf{O}ptimal \textbf{M}otion \textbf{P}lanner (\textbf{SCTOMP}): an \emph{offline} motion planning approach capable of \emph{computing dynamically feasible, collision-free and time-optimal trajectories}, by solely relying on a goal location and a geometric representation of the obstacle-free environment. 

For this purpose, we formulate a three-stage approach, where, firstly collision-free navigation corridors are found, secondly parametric paths within all corridors are computed, and thirdly a spatially reformulated time minimization per corridor is solved. In contrast to the aforementioned decoupled approaches, the parametric paths obtained in the second stage are not a geometric reference, but an extension of the free space associated with the respective corridor, and consequently, have no effect on the optimality of the time-minimizing problem in the third stage.

The presented scheme consists of three main ingredients: 1) Through the use of a spatial reformulation presented in \cite{arrizabalaga2022spatial}, we perform a spatial transformation of the time-based system states to path coordinates. The resultant system dynamics evolve according to a \emph{path parameter} $\xi$ instead of \emph{time} $t$. 2) We leverage Pythagorean Hodograph curves to efficiently and analytically compute the parametric functions required by this spatial reformulation. 3) Using the first and second ingredients, the time minimization problem is reduced into a finite horizon spatial problem with convex spatial bounds.

\noindent More specifically, we make the following contributions:
\begin{enumerate}
    \item We extend the applicability of the spatial reformulation introduced in \cite{arrizabalaga2022spatial}, by showing how it allows to conduct a singularity-free spatial transformation of the dynamics for any arbitrary system.
    \item We identify a computationally tractable methodology to compute Pythagorean Hodograph splines in a closed form, without the need for additional optimizations.
    \item We derive a motion planning methodology that, given a nonlinear system and constrained environment, finds the collision-free and dynamically feasible time-optimal trajectory.
\end{enumerate}

The remainder of this paper is structured as follows: Section~\ref{sec:problem} introduces the spatially constrained time-optimal motion planning problem. Section~\ref{sec:solution_approach} presents the solution proposed in this paper, by revisiting the aforementioned spatial reformulation, its compatibility with PH curves, and performing a spatial transformation of the time minimization problem. Experimental results are shown in Section~\ref{sec:experiments} before Section~\ref{sec:conclusion} presents the conclusions.

\vspace{2mm}
\noindent\textit{Notation:} We will use $\dot{(\cdot)} = \dv{(\cdot)}{t}$ for time derivatives and $(\cdot)' = \dv{(\cdot)}{\xi}$ for differentiating over path parameter $\xi$. For readability we will employ the abbreviation $ \txi=\xi(t)$.

\section{PROBLEM STATEMENT}\label{sec:problem}
\noindent We consider continuous time, nonlinear systems of the form
\begin{subequations}\label{eq:nonlinear_sys}
\begin{flalign}
    &\dot{\bm{x}}(t) = f(\bm{x}(t),\bm{u}(t)),\quad \bm{x}(t_0) = \bm{x_0}\,,\label{eq:nonlinear_sys1}\\
    &\bm{y}(t) = h(\bm{x}(t))\label{eq:nonlinear_sys2}\,,
\end{flalign}
\end{subequations}
where $\bm{x}\in\mathcal{X}\subseteq\mathbb{R}^{n_x}$ and $\bm{u}\in\mathcal{U}\subseteq\mathbb{R}^{n_u}$ define state and input constraints. Function $f\,:\,\mathbb{R}^{n_x}\times\mathbb{R}^m\mapsto\mathbb{R}^{n_x}$ refers to the equations of motion of an arbitrary dynamic system,  while the map $h\,:\,\mathbb{R}^{n_x}\mapsto\mathbb{R}^{n_y}$ defines the output of the system $\bm{y}\in\mathbb{R}^{n_y}$. Both $f$ and $h$ functions are assumed to be sufficiently continuously differentiable. The initial state is contained in a subset of the constrained state set, i.e.,  $\bm{x}_0\in\mathcal{X}_0\subseteq\mathcal{X}$.

To ensure that the system remains in the obstacle free space, the geometrically constrained environment is described as
\begin{equation}\label{eq:free_space}
    \Omega = \{g(h(\bm{x})) < 0\,,\quad\forall \bm{x}\in\mathcal{X}\}\, \in\mathbb{R}^{n_y}\,,
\end{equation}
where we also assume the function $g\,:\,\mathbb{R}^{n_y}\mapsto\mathbb{R}^{n_y}$ to be sufficiently continuously differentiable.

Spatially Constrained Time-Optimal Motion Planning refers to the problem of planning a trajectory that steers system~\eqref{eq:nonlinear_sys} from its initial state $\bm{x_0}$ to a goal output state $\bm{y}_\mathbf{f}\in\mathbb{R}^{n_y}$ in minimum-time, without compromising the system dynamics in~\eqref{eq:nonlinear_sys1}, ensuring the integrity of state and input constraints -- $\bm{x}\in\mathcal{X}$,  $\bm{u}\in\mathcal{U}$ -- and guaranteeing that output~\eqref{eq:nonlinear_sys2} remains within the free space in~\eqref{eq:free_space}, i.e., $\bm{y}\in\Omega$. This is equivalent to solving the following optimal control problem (OCP):
\begin{subequations}\label{eq:minT_cont}
    \begin{flalign}
    &\qquad\qquad\min_{\bm{x}(\cdot),\bm{u}(\cdot)} T = \int_0^{T} dt&
    \end{flalign}
    \vspace{-5mm}
	\begin{alignat}{3}
	\text{s.t.}\quad& \bm{x}(0) = \bm{x_0}\,,\\
	&\dot{\bm{x}} = f(\bm{x}(t),\bm{u}(t)), &\quad&t \in \left[0,T\right]\\
	&\bm{x}(t)\in\mathcal{X}\,,\,\bm{u}(t)\in\mathcal{U}\,,    &\quad&t \in \left[0,T\right]\\
	&\bm{y}(t)\in\Omega\,,    &\quad&t \in \left[0,T\right]\\
	&\bm{y}(T) = \bm{y}_\mathbf{f}\,.
	\end{alignat}

\end{subequations}
As mentioned earlier, \emph{decoupled} approaches separate this problem into a path planning and path tracking/following problem. The computational tractability advantages brought by these methods come at the expense of an inherited suboptimality in the planned trajectories. We seek to close this gap by formulating a method that \emph{directly} solves the minimum-time problem~\eqref{eq:minT_cont}, eliminating the need for the path planning stage, and thus leveraging the entire free space to compute the time-optimal trajectory.

\section{SOLUTION APPROACH}\label{sec:solution_approach}
\noindent The motion planning scheme presented in this paper leverages (i) a \emph{spatial transformation of the system dynamics} with respect to (ii) an \emph{arbitrary parametric curve with an associated adapted frame} to perform (iii) a \emph{spatial reformulation of the time-minimizing problem}~\eqref{eq:minT_cont}. These three ingredients are the main building blocks of the proposed methodology. In this section, we present further details on each of them.

\subsection{SPATIAL TRANSFORMATION OF SYSTEM DYNAMICS}

\noindent Let $\Gamma$ refer to a geometric reference and be defined as a path with an associated adapted-frame, whose position and orientation are given by two sufficiently continuous functions, $\bm{\gamma}\,:\,\mathbb{R}\mapsto\mathbb{R}^3$ and $\text{R}\,:\,\mathbb{R}\mapsto\mathbb{R}^{3\times3}$, that depend on path parameter $\xi$:
\begin{equation}\label{eq:path}
    \Gamma = \{\xi \in[\xi_0,\xi_f] \subseteq\mathbb{R}\mapsto\bm{\gamma}(\xi) \in \mathbb{R}^3, \text{R}(\xi) \in \mathbb{R}^{3\times3}\}
\end{equation}
Decomposing the adapted-frame into its components $\text{R}(\xi) = \{\bm{e_1}(\xi),\bm{e_2}(\xi),\bm{e_3}(\xi)\}$ allows to define its angular velocity vector as 
$\bm{\omega}(\xi) = \rchi_1(\xi)\bm{e_1}(\xi) + \rchi_2(\xi)\bm{e_2}(\xi) + \rchi_3(\xi)\bm{e_3}(\xi)$,
where $\rchi(\xi) = \{\rchi_1(\xi),\rchi_2(\xi),\rchi_3(\xi)\}\in\mathbb{R}^3\mapsto\mathbb{R}$ is also sufficiently continuous. 

\begin{figure}
	\centering
	\includegraphics[width=\linewidth]{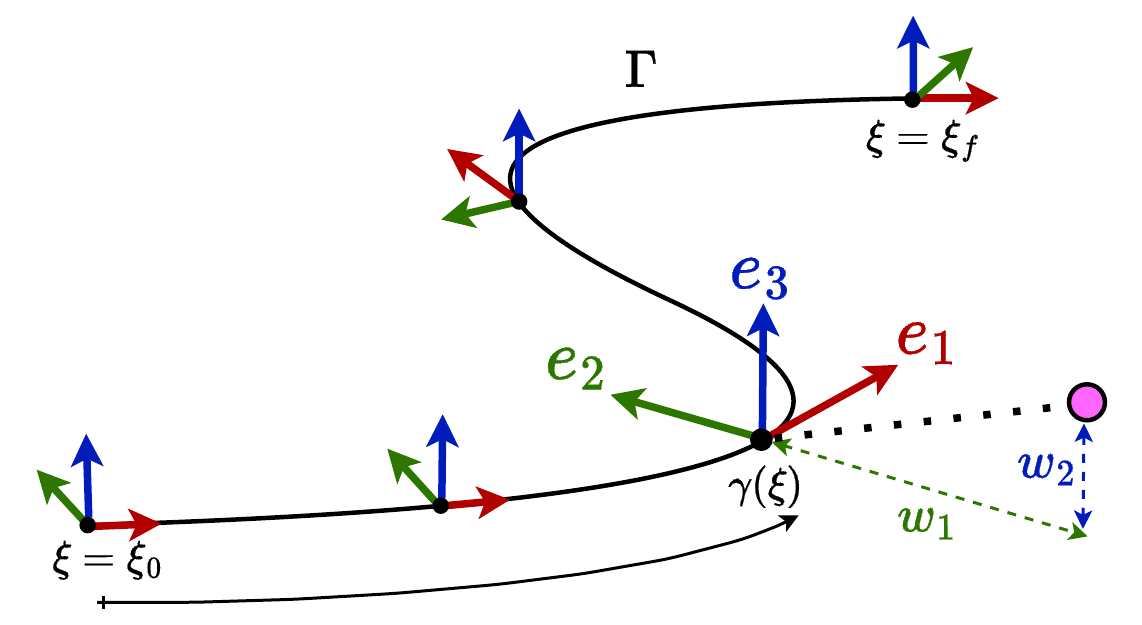}
	\caption{Spatial projection of the three-dimensional Euclidean coordinates, represented by the pink dot, onto a geometric path $\Gamma$ with an associated adapted-frame $\text{R}(\xi) = \{\bm{e_1}(\xi),\bm{e_2}(\xi),\bm{e_3}(\xi)\}$. The distance to the closest point on the path $\bm{\gamma}(\xi)$ is decomposed into the transverse coordinates $\bm{w} = \left[w_1,w_2\right]$.}\label{fig:spatial_coordinates}
\end{figure}

In \cite{arrizabalaga2022spatial} we demonstrated that the equations of motion of the spatial coordinates -- progress along the path $\xi$ and the orthogonal distance to it $\bm{w}=\left[w_1,w_2\right]$ -- associated to a point-mass moving at velocity $\bm{v}\in\mathbb{R}^3$ with respect to path $\Gamma$ are given by
\begin{subequations}\label{eq:point_mass_eq}
\begin{gather}
    \dot{\xi}(t) = \frac{\bm{e_1}(\txi)^\intercal \bm{v}(t)}{\sigma(\txi) - \rchi_3(\txi) w_1(t) + \rchi_2(\txi) w_2(t)}\,,\label{eq:xidot}\\
    \dot{w}_1(t) = \bm{e_2}(\txi)^\intercal \bm{v}(t) + \dot{\xi}(t) \rchi_1(\txi) w_2(t)\,,\\
    \dot{w}_2(t) = \bm{e_3}(\txi)^\intercal \bm{v}(t) - \dot{\xi}(t) \rchi_1(\txi) w_1(t)\, ,
\end{gather}
\end{subequations}
where $\sigma(\cdot)$ stands for the parametric speed of path $\Gamma$. For a better understanding of the spatial states, see Fig.~\ref{fig:spatial_coordinates}. These states, alongside their respective equations of motion~\eqref{eq:point_mass_eq}, allow for a projection of the Euclidean coordinates into the geometric path $\Gamma$, resulting in a change of coordinates. Doing so, embeds the geometric properties \emph{around path $\Gamma$} into the system dynamics, and thus, it has become a very popular approach among recent path following methods \cite{arrizabalaga2022towards,spedicato2017minimum,van2016path,kloeser2020nmpc}.

Given that our solution aims to be agnostic from a geometric reference, we perform a spatial transformation of the system dynamics in~\eqref{eq:nonlinear_sys}. To do so, we leverage the chain rule as follows:
\begin{gather}\label{eq:spatial_transform1}
    \bm{x}'\coloneqq \dv{\bm{x}}{\xi} = \dv{\bm{x}}{t}\dv{t}{\xi}
\end{gather}
Noticing that $\dv{t}{\xi} = 1/\dot{\xi}$ and assuming that $\dot{\xi}\neq0$, \eqref{eq:spatial_transform1} is simplified into
\begin{gather*}
    \bm{x}'= \frac{1}{\dot{\xi}} f(\bm{x},\bm{u})\,,\quad \forall \dot{\xi}\neq0\,.
\end{gather*}
The resultant spatially transformed equations of motion for the dynamic system~\eqref{eq:nonlinear_sys} are
\begin{subequations} \label{eq:nonlinear_sys_param}
\begin{flalign}
    &\bm{x}'(\xi) = \frac{f(\bm{x}(\xi),\bm{u}(\xi))}{\dot{\xi}(\bm{x}(\xi),\bm{u}(\xi))},\quad\label{eq:transformed_dynamics} \bm{x}(\xi_0) = \bm{x_0}\,,\\
    &\bm{y}(\xi) = h(\bm{x}(\xi))\,,
\end{flalign}
\end{subequations}
where $\dot{\xi}(\bm{x}(\xi),\bm{u}(\xi))$ is obtained from~\eqref{eq:xidot}. Comparing the \emph{spatially transformed} system~\eqref{eq:nonlinear_sys_param} with respect to the original system~\eqref{eq:nonlinear_sys}, it becomes apparent that the dynamics evolve with respect to \emph{path parameter $\xi$} instead of \emph{time $t$}. 

Transforming the system dynamics according to the chain rule~\eqref{eq:spatial_transform1} is a mature technique \cite{shiller1989robot,verscheure2009time}. Nevertheless, to the best of the authors' knowledge, this is the first time that it is applied to the recently derived spatial reformulation~\eqref{eq:point_mass_eq}. When doing so, the benefits of this reformulation are inherited by the transformed system~\eqref{eq:nonlinear_sys_param}. As opposed to state-of-the-art Frenet-Serret based reformulations \cite{van2016path,verschueren2016time,spedicato2017minimum}, \eqref{eq:point_mass_eq} does not take any assumptions in its adapted frame, and as result, it overcomes the two major drawbacks of the Frenet-Serret frame: (i) \emph{singularities} when the curvature vanishes and (ii) an \emph{undesired twist} with respect to its tangent component. 

\subsection{PYTHAGOREAN HODOGRAPH SPLINES}
\noindent For a complete definition of the spatially transformed system~\eqref{eq:nonlinear_sys_param}, the \emph{parametric functions} associated to $\xi$ -- the parametric speed $\sigma(\xi)$ and the adapted frame's rotation matrix $\text{R}(\xi)$ and angular velocity $\bm{\omega}({\xi})$ -- need to be expressed in closed form. To this end, we employ Pythagorean Hodograph (PH) curves.
\vspace{1mm}
\subsubsection{Preliminaries on PH curves}
\hspace{1cm}

\vspace{1mm}
\noindent PH curves are a subset of polynomial curves whose parametric speed is a polynomial of the path parameter $\xi$ \cite{farouki2008pythagorean}. Decomposing the components of $\bm{\gamma}(\xi)$ in~\eqref{eq:path} into $x(\xi),y(\xi),z(\xi)$, the condition for a polynomial to be a PH curve is equivalent to
\begin{gather}\label{eq:ph_cond}
    \sigma^2(\xi) = {x'}^2(\xi)+ {y'}^2(\xi) + {z'}^2(\xi)\,,
\end{gather}
where $\sigma(\xi)$ is a polynomial.  As proven in \cite{dietz1993algebraic}, every term in~\eqref{eq:ph_cond} can be expressed in terms of a \emph{quaternion polynomial} $\bm{Z}(\xi) = u(\xi) + v(\xi)\textbf{i} +g(\xi)\textbf{j} +h(\xi)\textbf{k}$ \footnote{$\{\textbf{i},\textbf{j},\textbf{k}\}$ refers to the $\mathbb{R}^{4}$ standard basis}, whose components $u(\xi),v(\xi),g(\xi),h(\xi)$ are also polynomial functions of the path parameter $\xi$. Consequently, the parametric speed can be reformulated as
\begin{equation}\label{eq:sigma_sq}
    \sigma(\xi)  = u^2(\xi) + v^2(\xi) + g^2(\xi) + h^2(\xi)\,.
\end{equation}
Another notable benefit of PH curves is that they inherit a continuous adapted frame, denoted as \emph{Euler Rodrigues Frame} (ERF), which also exclusively depends on its quaternion polynomial \cite{choi2002euler}.  Its respective rotation matrix is given by
\begin{equation}\label{eq:erf}
    \text{R}(\xi) = \frac{\left[\bm{Z}(\xi)\bi\bm{Z}^*(\xi), \bm{Z}(\xi)\bj\bm{Z}^*(\xi), \bm{Z}(\xi)\bz\bm{Z}^*(\xi)\right]}{|\bm{Z}(\xi)|^2}\,,
\end{equation}
where $(\cdot)^*$ refers to the quaternion's conjugate. Finally, considering that the components of its angular velocity are expressed as
\begin{equation}\label{eq:angvel_comp}
    \rchi(\xi) = \{\bm{e_2}'(\xi)\,\bm{e_3}(\xi),\bm{e_3}'(\xi)\,\bm{e_1}(\xi),\bm{e_1}'(\xi)\,\bm{e_2}(\xi)\}\,,
\end{equation}
the adapted frame's angular velocity  $\bm{\omega}(\xi)$ is likewise exclusively reliant on the quaternion polynomial. 

From~\eqref{eq:sigma_sq},~\eqref{eq:erf} and~\eqref{eq:angvel_comp}, it can be stated that all \emph{parametric functions} -- $\sigma(\xi),\text{R}(\xi),\bm{\omega}(\xi)$ -- solely depend on the quaternion polynomial $\bm{Z}(\xi)$, and consequently, they are fully defined by the coefficients of the underlying polynomial functions $u(\xi),v(\xi),g(\xi),h(\xi)$. For convenience we will refer to these as \emph{PH coefficients}, $\bm{\zeta}\in\mathbb{R}^{n_\zeta}$. Finally, notice that~\eqref{eq:sigma_sq} entails that a quaternion polynomial $\bm{Z}(\xi)$ of degree $n$ corresponds to a curve $\bm{\gamma}(\xi)$ of degree $2n+1$, while relying just on $4(n+1)$ PH coefficients. This implies that all three parametric functions, and hence the spatial reformulation underpinning system~\eqref{eq:nonlinear_sys_param}, are effectively encoded into a handful of coefficients.

\vspace{1mm}
\subsubsection{Efficiently computing collision-free PH splines}\label{subsec:eff_comp_ph}
\hspace{1cm}

\vspace{1mm}
\noindent To ensure planning capabilities within highly non-convex and complex environments, we concatenate multiple PH curves into a \emph{PH spline}. As mentioned before, the PH spline is not a geometric reference, but a parametric path that encodes the geometric properties of the environment into the spatially transformed system dynamics~\eqref{eq:nonlinear_sys_param}. Thus, as proven in the upcoming Section~\ref{sec:planar_example}, it has no influence on the time-optimality of the computed trajectory. However, when choosing the PH spline the requirements are twofold: First, the parametric functions -- $\sigma(\xi),\text{R}(\xi),\bm{\omega}(\xi)$ -- need to remain sufficiently often continuously differentiable, and second, it must be located within the obstacle-free space. 

For this purpose, \cite{arrizabalaga2022spatial} reduces the amount of free PH coefficients according to the continuity conditions, and subsequently, tailors the shape of the PH spline based on a desired criterion by solving an optimization problem on the remaining coefficients. To ensure the integrity of the spatial bounds it performs a convex decomposition of the free space $\Omega$ and constraints the control points of each segment of the PH spline to be encompassed inside the associated convex set. These control points can readily be obtained from a function that takes the starting location of the spline and the PH coefficients as inputs \cite{farouki2008pythagorean}. However, given that the coefficients relate to the hodograph, these constraints, as well as the cost function, are highly nonlinear and nonconvex, resulting in a large and numerically involved nonlinear program.

To alleviate this burden, rather than executing a convex decomposition and directly computing a PH spline, the proposed methodology in the \colorbox{stage0}{\textbf{first-stage}} \textcolor{stage0_dark}{identifies all navigation corridors within the free-space} $\bm{\mathcal{C}} = \{\mathcal{C}_1,...\,,\mathcal{C}_m\}\subseteq\Omega$, and in the \colorbox{stage1}{\textbf{second-stage}} \textcolor{stage1_dark}{locates an arbitrary curve}, enclosed within each of these corridors $\lambda_{1,\,...,m}$, \textcolor{stage1_dark}{and} \textcolor{stage1_dark}{transforms each of these curves into a PH spline} whose coefficients are $\bm{\mathcal{Z}}_{1,\,...,m}$. 

The presented hierarchical methodology exhibits modularity in that the finding of the corridors, computation of the curves and the successive conversion algorithm are entirely decoupled. Specifically, the former can be carried out effectively by employing state-of-the-art planning algorithms \cite{gonzalez2015review,liu2017planning,tordesillas2019faster}, whereas for the latter, we expand upon the $C^2$ hermite interpolation algorithm outlined in \cite{vsir2007} to $C^4$. Given that the conversion algorithm is presented as a closed-form solution, it entails no computational cost, and the burden of solving the aforementioned intractable nonlinear program is diminished to identifying a collision-free curve. 


\subsection{TIME MINIMIZATION: A SPATIAL APPROACH}
\noindent In order to encompass the entirety of the free space $\Omega$ and ensure time-optimality of the calculated trajectory, the \colorbox{stage2}{\textbf{third-stage}} of SCTOMP involves \textcolor{stage2_dark}{solving a time-minimizing OCP for all corridors} $k=1,,...,m$ and selecting the one with the minimum navigation time.

At a given corridor $k$, the PH coefficients $\bm{\mathcal{Z}}_k$ associated to the PH spline computed in the second-stage explicitly describe the parametric coefficients, and therefore, allow to rewrite the equation of motion of the spatially transformed system~\eqref{eq:transformed_dynamics} as
\begin{equation} \label{eq:nonlinear_sys_param_ph}
    \bm{x}'(\xi,\bm{\mathcal{Z}}_k) = \frac{f(\bm{x}(\xi),\bm{u}(\xi))}{\dot{\xi}(\bm{x}(\xi),\bm{u}(\xi),\bm{\mathcal{Z}}_k)}\,.
\end{equation}
This equation represents the spatially transformed dynamics model of a system whose temporal dynamics are given by $f$, with respect to a PH spline defined by PH coefficients $\bm{\mathcal{Z}}_k$. In contrast to~\eqref{eq:nonlinear_sys_param}, the system in~\eqref{eq:nonlinear_sys_param_ph} accounts for a method to compute the underlying parametric functions -- based on PH splines --, and thus, it is fully defined. 

\begin{figure*}[t]
\centering
\includegraphics[width=\textwidth]{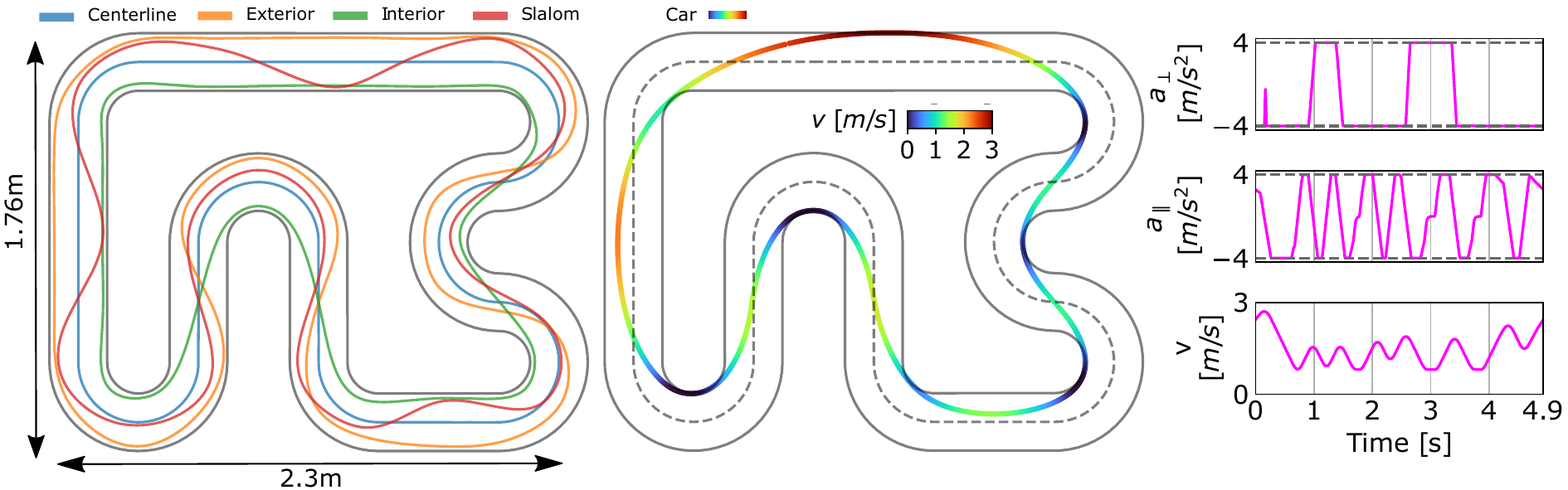}
\caption{Planar application of the proposed motion planning approach to a 1:43 scale autonomous car. The lines colored in light-gray represent the bounds of the race-track. \emph{Left:} The four PH splines used to demonstrate the invariance of the solution with respect to PH spline modifications. For clarity, their respective adapted frames are omitted. \emph{Center:} The resultant trajectory obtained for all four PH splines is given by the colored line. The color mapping relates to the car's longitudinal velocity. \emph{Right:} The first two rows represent the lateral and longitudinal accelerations, while the third shows the car's longitudinal velocity, all with respect to time.}\label{fig:car_results}
\vspace{-3mm}
\end{figure*}
\begin{table*}[t!]
    \renewcommand{\arraystretch}{1.25} 
	\centering
	\caption{States, inputs and constraints of the planar (car) and spatial (quadrotor) systems.} \label{tab:states_inputs_constraints}
	\begin{tabular}{|c||c|c|c|}
		\hline
		System & States $\bm{x}$ & Inputs $\bm{u}$ & Constraints \\
		\hline
		Car & $\{\xi,w,\psi,v,D,\delta\}$ & $\{\dot{D},\dot{\delta}\}$ & $D\in[-1,1]$\,,\;$\delta\in[-0.4,0.4]$\,,\;$\dot{D}\in[-10,10]$\,,\;$\dot{\delta}\in[-2,2]$\,,\;$\{a_\parallel, a_\perp\}\in[-4,4]$\\
		\hline
		Quadrotor  & $\{\xi,w_1,w_2,\bm{v},\bm{q}\}$ & $\{f_c,\bm{\omega}\}$& $f_c\in[0,27.52]$\,,\;$\{\omega_x,\omega_y\}\in[-15,15]$\,,\;$\omega_z\in[-0.3,0.3]$ \\
		\hline
	\end{tabular}
\vspace{-5mm}
\end{table*}
Revisiting the fact that  $\dv{t}{\xi} = 1/\dot{\xi}$ and leveraging the spatially transformed system dynamics in~\eqref{eq:nonlinear_sys_param_ph}, the time-minimizing problem within the free space associated to corridor $k$ can be reformulated as
\begin{subequations}\label{eq:minT_par}
    \begin{flalign}
     &\qquad\qquad\min_{\bm{x}(\cdot),\bm{u}(\cdot)} T_k = \int_{\xi_0}^{\xi_f}\frac{1}{\dot{\xi}(\bm{x}(\xi),\bm{\mathcal{Z}}_k)}\, d\xi&
    \end{flalign}
    \vspace{-5mm}
	\begin{alignat}{3}
	\text{s.t.}\quad& \bm{x}(\xi_0) = \bm{x_0}\,,\\
	&\bm{x}' = \frac{f(\bm{x}(\xi),\bm{u}(\xi))}{\dot{\xi}(\bm{x}(\xi),\bm{\mathcal{Z}}_k)}, &\quad&\xi \in \left[\xi_0,\xi_f\right]\label{eq:dynamic_const}\\
	&\bm{x}(\xi)\in\mathcal{X}\,,\,\bm{u}(\xi)\in\mathcal{U}\,,    &\quad&\xi \in \left[\xi_0,\xi_f\right]\\
	&\bm{y}(\xi)\in \mathcal{C}_k\,, \label{eq:spatial_constr}   \\
	&\bm{y}(\xi_f) = \bm{y}_\mathbf{f}\,.
 \label{eq:mpc_spatial_cond_cont}
	\end{alignat}
 \noindent The resultant OCP~\eqref{eq:minT_par} is a finite horizon problem, and unlike the original problem~\eqref{eq:minT_cont}, the integration interval on which we solve the optimization is independent of the decision variables. Similar spatial reformulations of the time minimization problem have already been conducted previously \cite{verschueren2016time, spedicato2017minimum}. However, these are system-specific and inherit the limitations to Frenet-Serret based reformulations. Both shortcomings are solved by \eqref{eq:minT_par}, which is compatible with any system whose dynamics are given by $f$ and applicable to a geometric path whose adapted frame can be tailored by PH coefficients $\bm{\mathcal{Z}}_k$. Besides that, in a similar manner to \cite{arrizabalaga2022towards,spedicato2017minimum}, combining the spatial transformation in~\eqref{eq:nonlinear_sys_param_ph} with the change of coordinates from the Euclidean position to the spatial coordinates, i.e., $[p_x,p_y,p_z]\rightarrow[\xi,w_1,w_2]$, allows to reformulate the spatial bounds in~\eqref{eq:spatial_constr} as convex constraints. This can be tailored to the corridor's definition, e.g., either as $||\bm{w}||_2^2 < \delta(\xi)^2$ or $A(\xi) \bm{w} - b(\xi) \leq 0$,  where $\bm{w}=[w_1,w_2]$, and $\{A(\xi), b(\xi)\}$, $\delta(\xi)$ are the halfspace representation and the radius of the corridor's cross section at $\xi$, respectively.
 

\end{subequations}



A summary of the proposed motion planning scheme, showing the \textcolor{stage0_dark}{first}, \textcolor{stage1_dark}{second}- and \textcolor{stage2_dark}{third}-stages is depicted in Algorithm~\ref{alg:algorithm1}. 

\begin{algorithm}[h]
\caption{\textit{Spatially Constrained Time-Optimal Motion Planning} (\textbf{SCTOMP}): Given state $\bm{x_0}$, goal $\bm{y}_\mathbf{f}$ and environment $\Omega$, find the time-optimal set of states and inputs $\bm{x}^*,\bm{u}^*$}\label{alg:algorithm1}
\begin{algorithmic}[1]
\Function{SCTOMP}{$\bm{x_0}$, $\bm{y}_\mathbf{f}$, $\Omega$}
\State $\bm{\mathcal{C}} =\mathcal{C}_1,...\,,\mathcal{C}_m \gets \textsc{\textcolor{stage0_dark}{Corridors}}(\bm{x_0},\bm{y}_\mathbf{f},\Omega)$~\cite{gonzalez2015review}
\State $T^* \gets \infty$
\For{$k\in\{1,\,...,m\}$}
\State $\lambda_k \gets \textsc{\textcolor{stage1_dark}{Find Collision-Free Path}}(\bm{x_0},\bm{y}_\mathbf{f},\mathcal{C}_k)$
\State $\bm{\mathcal{Z}}_k \gets \textsc{\textcolor{stage1_dark}{Convert to PH Spline}}(\lambda_k)$~\cite{vsir2007}
\State $\bm{x}_k,\bm{u}_k, T_k \gets \textsc{\textcolor{stage2_dark}{Time Min.}}(\bm{x_0},\bm{y}_\mathbf{f},\mathcal{C}_k,\bm{\mathcal{Z}}_k)$~\eqref{eq:minT_par}
\If{$T_k<T^*$}
    \State $\bm{x}^*,\bm{u}^* \gets \bm{x}_k,\bm{u}_k$
\EndIf
\EndFor
\State\Return $\bm{x}^*,\bm{u}^*$
\EndFunction
\end{algorithmic}
\end{algorithm}
\section{EXPERIMENTS}\label{sec:experiments}
\noindent To evaluate our approach, we divide the experimental analysis into two parts, including systems with different dynamics. First, we test SCTOMP by analyzing its performance in a well-known single-corridor planar scenario (2D, Sec.~\ref{sec:planar_example}), demonstrating that the proposed scheme remains time-optimal regardless of the underlying PH spline. Secondly, the evaluation is expanded to a spatial system (3D, Sec.~\ref{sec:spatial_example}) and compared against existing state-of-the-art approaches in a range of case-studies across two different cluttered environments.

\vspace{1mm}
\noindent \emph{Numerical implementation:} The OCP~\eqref{eq:minT_par} is modeled in CasADi \cite{andersson2019casadi} and solved by IPOPT \cite{wachter2006implementation} after being approximated by a multiple-shooting approach \cite{bock1984multiple}, where the optimization horizon $\xi_f-\xi_0$ is discretized into sections with constant decision variables. The integration routines respective to the dynamic constraints~\eqref{eq:dynamic_const} are approximated by a 4th-order Runge-Kutta method.
\subsection{PLANAR SYSTEM IN SINGLE-CORRIDOR SCENARIO}\label{sec:planar_example}
\noindent We assess the performance of our method in a state-of-the-art race track for 1:43 scale autonomous cars. Given that this case-study has been the research focus of previous work \cite{kloeser2020nmpc,verschueren2014towards}, its time-optimality is well-understood.

It is worth highlighting that our method shows its full potential in unstructured environments, where imposing a geometric reference might be detrimental for planning time-optimal trajectories. However, race-alike scenarios, consist of well-structured -- single-corridor-- environments in which the progress along the track and its boundaries are perfectly referenced by the centerline, and thus are better suited to path following methods, such as \cite{arrizabalaga2022spatial,romero2022model,kloeser2020nmpc}. Nevertheless, the proposed case-study allows for (i) performing an \emph{approximated benchmark} on time-optimality, (ii) demonstrating that the underlying PH splines have no effect on the computed trajectory  and (iii) showing the method's applicability to constrained planar systems.

To this end, in a similar way to \cite{kloeser2020nmpc} and \cite{verschueren2014towards}, we employ the dynamic bicycle model with states $\bm{x}=\left[p_x,p_y,\psi,v,D,\delta\right]\in\mathbb{R}^6$ and inputs $\bm{u}=\left[\dot{D},\dot{\delta}\right]\in\mathbb{R}^2$, where $\{p_x,p_y,\psi,v,D,\delta\}\in\mathbb{R}$, refer to the position, yaw, longitudinal velocity, throttle and steering angle. For the respective equations of motion and model coefficients, please refer to eq. 3 in \cite{kloeser2020nmpc}. To exploit the aforementioned convexity in the spatial constraints~\eqref{eq:spatial_constr}, we adopt a similar approach presented in \cite{spedicato2017minimum} and  \cite{arrizabalaga2022towards}, where the Euclidean position coordinates $[p_x,p_y]$ are projected onto their corresponding spatial counterparts $[\xi,w]$. In addition, to guarantee the validity of the first-principles model's approximation, we impose constraints on the throttle, steering angle, their respective change rates, as well as the longitudinal and lateral accelerations $\{a_\parallel,a_\perp\}\in\mathbb{R}$. The numerical values associated with these constraints can be found in Table~\ref{tab:states_inputs_constraints}.

Following Algorithm~\ref{alg:algorithm1}, and given the single-corridor nature of the present case-study, the first-stage results in the race-track itself. Subsequently,
before solving the time-minimizing problem, in the second-stage we need to compute a PH spline that lies within the race-track. As discussed in Section~\ref{subsec:eff_comp_ph}, the definition of the collision-free curve upon which the spline is converted does not influence the computation of the time-optimal trajectory. Putting it another way, any curve that lies within the race-track is eligible to be embedded as a PH spline into the time optimization problem~\eqref{eq:minT_par}. Nonetheless, because the obstacle-free space remains fixed, the time-optimal solution is invariant regardless of the PH spline used.
\begin{table}[t!]
    \renewcommand{\arraystretch}{1.25} 
	\centering
	\caption{A comparison of the four cases employed for analyzing the sensitivity of the time-optimal solution with respect to PH spline variations. The features of the resultant splines -- curvature energy and arc-length -- are given in the second and third column, while the respective navigation times -- computed from~\eqref{eq:minT_par} -- are shown in the fourth column. All four PH splines, as well as the equivalent trajectory obtained for all the splines are shown in Fig.~\ref{fig:car_results}.} \label{tab:ph_spline_diff}
	\begin{tabular}{|c||c|c|c|}
		\hline
		PH spline & Energy [-] & Arc-length [m] & Time [s]\\
		\hline
		Centerline  & 39.16 & 8.71 & 4.924\\
		\hline
		Exterior & 35.65 & 9.98 & 4.920\\
		\hline
		Interior & 49.50 & 7.25 & 4.925\\
		\hline
  		Slalom & 46.16 & 9.25 & 4.923\\
		\hline
	\end{tabular}
\vspace{2mm}
\end{table}
To demonstrate the non-sensitivity of SCTOMP with respect to PH splines, in the second-stage we compute four different versions, according to four different criteria: the track's center, outer and inner-lines, as well as an intermediary slalom. Their geometrical locations are visualized in the left of Fig.~\ref{fig:car_results} and their characteristics are given in the second and third columns of Table~\ref{tab:ph_spline_diff}. Comparing the PH splines against each other, the one denoted as \emph{interior} is the shortest and sharpest, followed by \emph{centerline}, while \emph{exterior} and \emph{slalom} are smoother, longer and more curvy. Their differences on shape and size allow for challenging our methodology's robustness with respect to PH spline variations. 

Upon solving the third-stage for all four splines, a closely-matched series of lap times are obtained, exhibiting an average duration of $4.922\pm0.002$\SI{}{\second} and a maximum gap of \SI{0.005}{\second}. These timings are presented in the fourth column of Table~\ref{tab:ph_spline_diff} and the minor differences in time are attributed to numerical implementation. In fact, all four solutions result in the same trajectory as the one stated in \cite{kloeser2020nmpc}, thereby rendering the aforementioned differences negligible and demonstrating that the obtained solution is agnostic of the underlying PH spline. 

The resultant trajectory is illustrated in the center column of Fig.~\ref{fig:car_results}. As common in racing scenarios, the time-optimal trajectory enters a corner from its inner side and exists from the outside. Moreover, the longitudinal and angular accelerations attached in the right column of Fig.~\ref{fig:car_results} show that the car's actuation is fully exploited by always remaining on its physical limit.
\subsection{SPATIAL SYSTEM IN MULTI-CORRIDOR SCENARIOS}\label{sec:spatial_example}
\noindent After analyzing SCTOMP's ability to compute time-optimal trajectories in a planar case-study with a single-corridor, we study its applicability to spatial systems within more challenging, i.e., multi-corridor, scenarios.

To this end, we focus on a benchmark introduced in~\cite{penicka2022minimum}, and further extended in~\cite{penicka2022learning}, which entails the navigation of a quadrotor between start and goal states in two intricate and congested environments: a "forest" characterized by scattered columns, and an indoor "office". To compute time-optimal trajectories for a quadrotor, it is necessary to select single rotor thrust commands as control inputs, as this approach enables full exploitation of the system's physical limits. However, to ensure a fair comparison, we adopt the same input modality as in \cite{penicka2022minimum,penicka2022learning}, namely, collective thrust and body rates. It is worth noting that this is the preferred control choice for professional human pilots and has recently been demonstrated to be the most successful input selection for training policies on quadrotors \cite{kaufmann2022benchmark}.

The states and inputs of the quadrotor are denoted as $\bm{x}=\left[\bm{p},\bm{v},\bm{q}\right]\in\mathbb{R}^{10}$, $\bm{u}=\left[f_c,\bm{\omega}\right]\in\mathbb{R}^4$, where $\{\bm{p},\bm{v}, \bm{\omega}\}\in\mathbb{R}^3$ refer to the position, velocity and body rates, $\bm{q}\in\mathbb{R}^4$ is the unit quaternion and $f_c\in\mathbb{R}$ is the collective thrust. The equations of motion of this system can be found in \cite{arrizabalaga2022towards}. For the specific coefficients we use the same racing quadrotor as in \cite{foehn2021time,penicka2022minimum,penicka2022learning}. As conducted in the preceding planar study, to capitalize the convexity of the spatial constraints~\eqref{eq:spatial_constr}, we perform a spatial reformulation from the Euclidean coordinates $\bm{p}$ to the spatial ones $[\xi,w_1,w_2]$. Moreover, to ensure the integrity of the first-principles model, the collective thrust and the angular velocities are bounded to the values in Table~\ref{tab:states_inputs_constraints}.

\begin{figure*}[t]
\centering
\includegraphics[width=\textwidth]{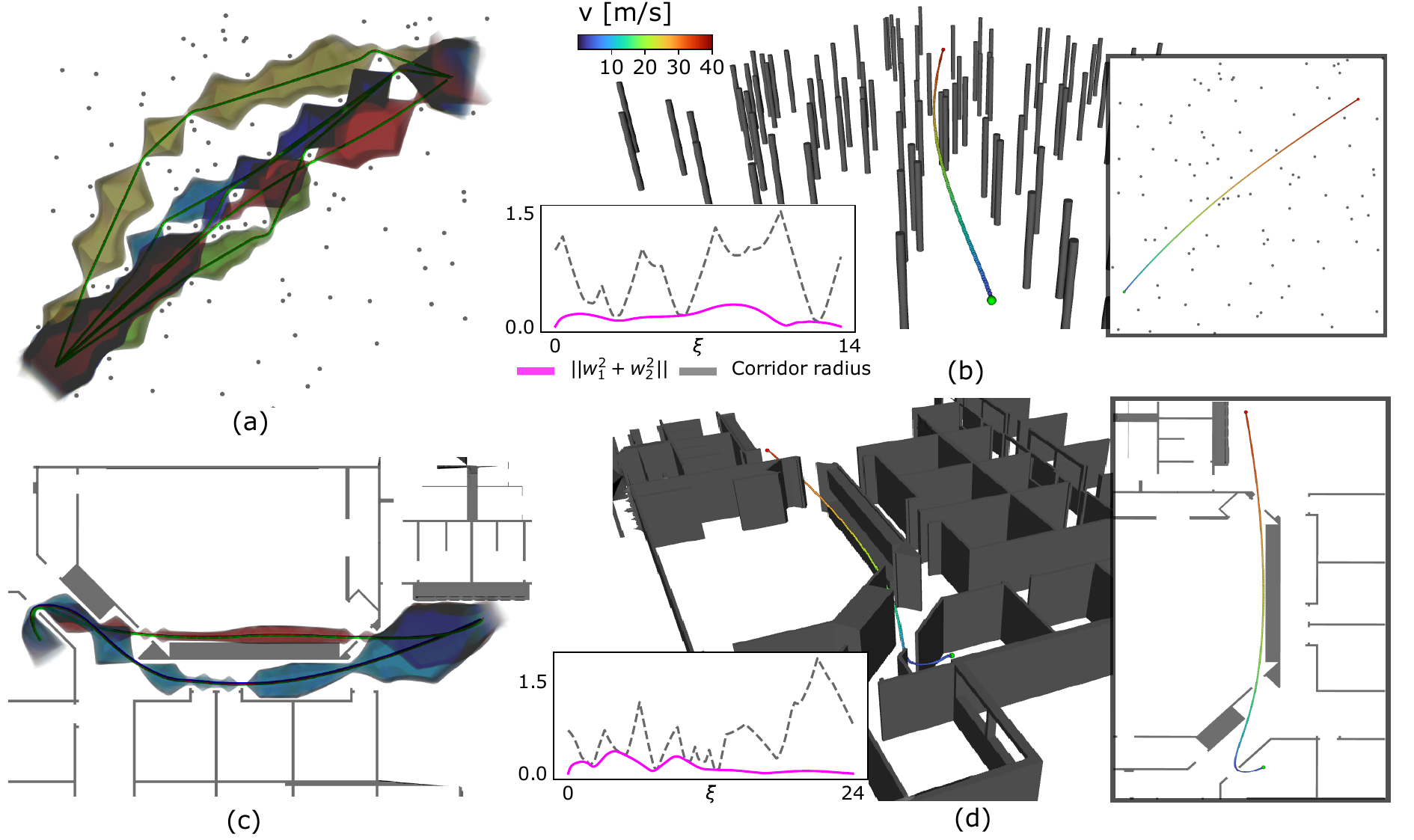}
\caption{Two 3D case-studies conducted to compare SCTOMP to baseline algorithms, with the numerical results presented in Table~\ref{tab:spatial_results}. The top row (a-b) pertains to test-case 3 in a forest environment, while the bottom row (c-d) corresponds to test-case 0 in an office environment. The left column (a, c) displays a top-view of the various corridors obtained from the first-stage, each of which is represented by the adapted frames respective to the underlying PH splines. The right column (b, d) shows the minimum-time trajectory of the quadrotor, denoted by a colored mapping that is indicative of the velocity norm. At the lower-left side, we plot the distance between the corridor's PH spline and the quadrotor --the norm of the transverse coordinates $w_1$ and $w_2$--, alongside the radius of the corridor, showcasing how the computed trajectory fully exploits the available space by tangentially touching its borders.}\label{fig:results}
\vspace{-7mm}
\end{figure*}

\begin{table}[b!]
    \renewcommand{\arraystretch}{1.25} 
	\centering
    \vspace{2.5mm}
	\caption{A comparison between the navigation times computed by baseline algorithms and our proposed method. The sampling-based timings show the average of 30 runs and the best in parenthesis.} \label{tab:spatial_results}
	\begin{tabular}{c c c c c c}
		\hline
  		Envir- & Test & CPC & Sampling- & RL & \textbf{SCTOMP}\\
		onment & case & \cite{foehn2021time} & based \cite{penicka2022minimum} & \cite{penicka2022learning} & \textbf{(Ours)}\\	
        \hline
		\multirow{3}{4em}{Forest} & 
          0 & 0.95 & 1.10$\,\pm\,$0.13 (0.96) & 0.98 & \textbf{0.94}\\
        & 2 & 0.95 & 0.98$\,\pm\,$0.01 (0.96) & 1.00 & \textbf{0.94}\\
        & 3 & - & 1.50$\,\pm\,$0.17 (1.30) & 1.28& \textbf{1.19}\\
  		\hline
		\multirow{4}{4em}{Office} & 
          0 & - & 2.38$\,\pm\,$0.28 (1.93) & 1.62 & \textbf{1.53}\\
        & 1 & - & 1.74$\,\pm\,$0.06 (1.69) & 1.64 & \textbf{1.54}\\
        & 2 & - & 2.20$\,\pm\,$0.13 (1.93) & \textbf{1.56} & 1.63\\
        & 3 & - & 1.81$\,\pm\,$0.11 (1.58) & 1.40 & \textbf{1.32}\\
		\hline
	\end{tabular}
\end{table}

We compare the performance of SCTOMP against three baseline algorithms. Among these, the first one is a variant of CPC \cite{foehn2021time}, a methodology capable of computing theoretically time-optimal trajectories according to the physical limits of the quadrotor, extended by the authors in~\cite{penicka2022minimum} to make it applicable in cluttered environments. The second baseline is the hierarchical sampling-based method~\cite{penicka2022minimum}, while the third refers to the deep Reinforcement Learning approach~\cite{penicka2022learning}. Each of these methods is tested in three and four case-studies, i.e., different start and goal positions in the forest and office environments, respectively. Notice that these case-studies are identical to the ones in~\cite{penicka2022minimum,penicka2022learning} and are listed in the second column of Table~\ref{tab:spatial_results}.

Leveraging the aforementioned modularity of the first-stage, we find the collision-free corridors by defining a tunnel around the topological paths computed in~\cite{penicka2022minimum}. The radius of the tunnel is specified by the distance to the closest point in an inflated voxel grid of the occupancy map, resulting in overly conservative corridors. As already mentioned, this stage is modular in such a way that it could be replaced by other methods, e.g., \cite{liu2017planning, tordesillas2019faster}. After computing all corridors and solving the second- and third-stages for each of them, the trajectory with lowest navigation time is picked. The corridors and trajectories computed for two different case-studies can be visualized in Fig~\ref{fig:results}. 

The obtained navigation times are reported in Table~\ref{tab:spatial_results}. Due to the randomness associated to the sampling-based method, we give the average duration for 30 runs, alongside the best solution in parentheses. The results show that SCTOMP is able to compute the trajectories with the minimum navigation time for all test-cases, except for one. As mentioned in~\cite{penicka2022minimum}, the extended CPC method only finds solutions for the easiest forest case-studies. Reductions of SCTOMP with respect to these theoretical lower bounds are attributed to avoiding the singularity $\dot{\xi}=0$; instead of starting from a stationary position, we initialize the quadrotor with a minimal velocity $||\bm{v}||_2\neq\,$\SI{0}{\meter\per\second}. In the remaining more complex test cases, SCTOMP shows capable of computing shorter time solutions than the baseline methods. On the one hand, the excessively hierarchical procedure of the sampling-based algorithm jeopardizes its time-optimality. On the other hand, aiming to account for tracking disturbances, the RL method learns to prioritize safety over time-optimality by avoiding to fly too close to obstacles, and thus, rendering the obtained trajectories sub-optimal. In contrast, SCTOMP fully exploits the free-space within a given corridor, resulting in trajectories that tangentially touch its borders. This feature can be visualized at the graphs attached to the lower-left side of Figs~\ref{fig:results}b and \ref{fig:results}d. As a consequence, once the corridor associated to the optimal trajectory is found, SCTOMP guarantees to compute the time-optimal trajectory. This relates to the second office case-study, where SCTOMP shows higher navigation times than the RL due to the fact that none of the corridors obtained in first-stage contained the optimal trajectory. This can be overcome by implementing more detailed sampling-based method in stage 1.


\section{CONCLUSION}\label{sec:conclusion}
\noindent In this work, we presented a motion planning approach capable of computing time-optimal trajectories in spatially constrained environments. The time-optimal trajectories obtained by our method, not only exploit the system's actuation, but also the available free space. For this purpose, we rely on a spatial reformulation that allows for performing a singularity-free spatial transformation of the system dynamics, as well as the time minimization problem. To compute the underlying parametric functions required by this reformulation, we leverage Pythagorean Hodograph splines. This results in a three-stage scheme, where after finding collision-free corridors with PH splines enclosed in them, their respective coefficients are fed into a spatially reformulated time minimization problem. Experiments on a single-corridor planar case-study show that the solution's time-optimality is agnostic to the underlying PH spline. Moreover, an extensive benchmark against baseline algorithms account for the method's time-optimal capabilities. Lastly, the versatility of the systems  -- a planar bicycle model and a spatial quadrotor -- and differences in the scale of the environments upon which the experiments were conducted emphasize the motion planner's generality and applicability to a wide range of systems and scenarios.
\newcommand{\BIBdecl}{\setlength{\itemsep}{0.085 em}} 
\bibliographystyle{IEEEtran}
\bibliography{spatially_constrained_time_optimal_motion_planning}

\end{document}